\documentclass{article}

\usepackage[dblblindworkshop, final]{neurips_2025}

\usepackage[utf8]{inputenc} 
\usepackage{newunicodechar}
\newunicodechar{−}{\ensuremath{-}}
\usepackage[T1]{fontenc}    
\usepackage{hyperref}       
\usepackage{url}            
\usepackage{booktabs}       
\usepackage{amsfonts}       
\usepackage{nicefrac}       
\usepackage{microtype}      
\usepackage{xcolor}         
\usepackage{graphicx}
\usepackage{subcaption}
\workshoptitle{Efficient Reasoning}

\title{Mechanistic Interpretability of GPT-2: Lexical and Contextual Layers in Sentiment Analysis
}

\author{
  Amartya Hatua \thanks{Code and data available at: \url{https://github.com/amartyahatua/MI_Sentiment_Analysis}}\\
  AI Center of Excellence\\
  Fidelity Investments\\
  Boston, MA 02210 \\
  \texttt{amartyahatua@gmail.com} \\
}

\begin{document}

\maketitle

\begin{abstract}
 We present a mechanistic interpretability study of GPT-2 that causally examines how sentiment information is processed across its transformer layers. Using systematic activation patching across all 12 layers, we test the hypothesized two-stage sentiment architecture comprising early lexical detection and mid-layer contextual integration. Our experiments confirm that early layers (0-3) act as lexical sentiment detectors, encoding stable, position specific polarity signals that are largely independent of context. However, all three contextual integration hypotheses: Middle Layer Concentration, Phenomenon Specificity, and Distributed Processing are falsified. Instead of mid-layer specialization, we find that contextual phenomena such as negation, sarcasm, domain shifts  etc. are integrated primarily in late layers (8-11) through a unified, non-modular mechanism. These experimental findings provide causal evidence that GPT-2's sentiment computation differs from the predicted hierarchical pattern, highlighting the need for further empirical characterization of contextual integration in large language models.
\end{abstract}

\section{Introduction}
Large language models demonstrate impressive capabilities across a wide range of diverse linguistic tasks. Despite this progress, existing interpretability research primarily relies on correlational evidence from probing or attention analysis. Consequently, the internal causal structure through which these models encode and transform linguistic information has not been widely explored. 

Early research focused on identifying how distinct layers within transformers contribute to different stages of linguistic processing. \cite{tenney2019bert} found that BERT processes language in stages early layers handle syntactic information, while later layers understand semantic relationships. This suggests that transformers operate similarly to a pipeline, progressing from simple features to a complex understanding. It was the first clear evidence that these models have organized, step by step processing. Building upon this foundation,  \cite{jawahar2019bertology}, formalized a three-tier hierarchical framework: early layers handle basic word features, middle layers deal with grammar and sentence structure, and late layers understand meaning and how distant words relate to each other. Simultaneously,  \cite{clark2019bert} revealed that individual heads develop specialized functions for specific linguistic phenomena, following the same early to late progression. In \cite{rogers2020primer}, a comprehensive synthesis was provided that established a general consensus on middle layer specialization for syntactic structure, while highlighting that semantic processing remains more distributed and less well understood. All these studies showed that transformers seem to process language in organized, step-by-step ways. However, their methodologies were predominantly correlational, relying on probing classifiers and attention analysis to identify what information exists in representations rather than what models actually use during inference.

Newer research has highlighted this gap. Scientists now realize that finding patterns doesn't prove the model actually uses them. As \cite{belinkov2023gap} puts it, there's a gap between what we can detect in the model and what the model actually relies on; just because we can find information doesn't mean the model uses it. When \cite{elazar2018adversarial} tried removing features they thought were important, the models often worked just fine without them. This suggested they were finding fake patterns, not real ones.  \cite{makelov2023interpretability} found ``interpretability illusions'' interventions that seemed to reveal how models work but were actually triggering backup systems that had nothing to do with normal processing. 

Recent years have brought major improvements in solving the correlation-causation problem. The field of mechanistic interpretability \cite{rai2024practical} has developed new techniques like activation patching \cite{heimersheim2024activation} that let researchers directly test cause and effect, while automated tools have made the analysis process more systematic. Companies like Anthropic and OpenAI have successfully applied these mechanistic methods  to real models, finding millions of interpretable features in their large language models. Among these challenging phenomena, sentiment analysis presents a particularly instructive case. Sentiment analysis represents a particularly complex challenge for mechanistic interpretability. Unlike syntactic phenomena that localize to specific layers or attention heads, sentiment processing requires sophisticated integration of lexical, syntactic, and contextual information. The exact words can convey opposite meanings depending on context—``great'' in ``great movie'' versus ``oh great, it's raining''—requiring dynamic contextual reasoning that current methods struggle to explain mechanistically.

In sentiment analysis, it remains essential to determine whether transformer layers interact in a truly causal manner. Do early layers construct representations that later layers build upon, or do they operate in parallel? Is contextual information processed locally or distributed across the network? Correlational analyses cannot resolve these questions. In this work, we employ activation patching and causal interventions to empirically examine sentiment processing in GPT-2, providing mechanistic evidence for stage wise organization and establishing a foundation for deeper causal interpretability research.

\section{Related Research}

\subsection{Transformer Layer Analysis and Interpretability}
The study of how transformers process language has advanced through several methodological stages. Early probing work \cite{tenney2019bert} showed that BERT encodes linguistic features hierarchically, with parts of speech in early layers, syntax in middle layers, and semantics in later layers—forming the basis of the layer specialization hypothesis.  \cite{jawahar2019bertology} expanded this view, showing a progression from surface features to long-distance dependencies, while  \cite{clark2019bert} demonstrated that attention heads develop specialized functions aligned with linguistic phenomena.  \cite{rogers2020primer} synthesized these findings in their BERTology survey, framing transformer processing as a three-stage pipeline of feature extraction, syntax, and semantics—a framework that remains influential but largely correlational.

\subsection{Methodological Evolution in Mechanistic Interpretability}
Limitations of correlational methods led to causal approaches in mechanistic interpretability. \cite{elhage2021mathematical} emphasized the need to distinguish between what information exists in representations and what computations models actually perform. Activation patching became central here, with \cite{wang2022indirect} showing distributed processing for indirect object identification and  \cite{meng2022locating} revealing that factual knowledge depends on interactions across layers rather than localized storage. Automated methods such as ACDC (\cite{conmy2023acdc}) extended this work by identifying candidate circuits, though many fail causal tests, highlighting challenges of functional faithfulness.

\subsection{Sentiment Processing in Language Models}
While sentiment analysis is a well-studied application of transformers, the mechanisms behind sentiment processing remain poorly understood. Probing studies (e.g., \cite{liu2019roberta}) showed that different layers encode aspects of sentiment, suggesting a multi-stage progression from word recognition to contextual modulation and final decision-making. Yet these findings are largely correlational, and sentiment’s contextual nature—shaped by negation, intensification, and pragmatic cues like sarcasm—complicates interpretation. Recent studies have examined such phenomena individually, but little work has addressed how they interact within a unified computational framework.

\subsection{Limitations of Current Approaches}
Current interpretability methods struggle with complex semantic phenomena like sentiment processing. Most mechanistic successes have been limited to simple tasks such as arithmetic or syntax, leaving contextual reasoning and pragmatic inference opaque. Probing approaches, though influential, often fail to capture true model behavior—high accuracy can reflect memorization rather than genuine computation ( \cite{hewitt2019designing}, \cite{belinkov2023gap}). Theories like layer specialization remain largely correlational, lacking causal validation, which is especially problematic for sentiment, where meaning depends on intricate contextual interactions.

\subsection{The Need for Systematic Causal Validation}
These limitations point to the need for systematic causal validation of transformer processing theories. New methods—such as causal scrubbing, refined activation patching, and sparse autoencoders—offer powerful tools but have rarely been applied to foundational questions. In sentiment analysis, we still lack a causal account of how the processing stages interact. This work provides the first systematic validation of the three-stage sentiment processing hypothesis, moving beyond correlation to genuine mechanistic insight.

\section{Methodology}

\subsection{Experimental Design}
We have tested the two-stage sentiment processing hypothesis in GPT-2 using activation patching across the 12 layers. The analysis focuses on lexical detection and contextual integration, with controlled interventions isolating the causal role of each stage in sentiment behavior. We use GPT-2 (117M) through TransformerLens \cite{nanda2023transformerlens}, which allows standardized access to activations and precise interventions. The model is run in inference mode without finetuning, with consistent tokenization and identical architecture across all conditions.

\subsection{Activation Patching Protocol}
Activation patching is used in this study as a causal intervention technique to identify which transformer layers directly contribute to sentiment processing. By selectively substituting internal activations between contrasting input sentences, we isolate the specific layers responsible for lexical detection and contextual integration. For each test pair, we conducted activation patching, replacing activations from the source sentence (e.g., positive sentiment) with those from the target sentence (e.g., negative sentiment) at each layer independently. The resulting change in sentiment classification probability was measured to quantify the causal contribution of each layer, where larger shifts indicate greater causal importance for sentiment processing.

\subsection{Lexical Detection}

We perform a linear probe on GPT-2's final layer representations to classify sentiment polarity, achieving 95\% accuracy on a held-out validation set. This probe serves as our behavioral measure for sentiment classification performance, allowing us to quantify how interventions affect the model's sentiment processing.

\subsection{Hypotheses}
We test four specific hypotheses about lexical processing:

\begin{enumerate}
    \item Lexical Sensitivity: Sentiment word substitutions produce measurable activation differences. 
    \item Early Layer Dominance: Layers 0-3 show the strongest effects for lexical sentiment. 
    \item Position Specificity: The effects concentrate on the positions of the sentiment words.
    \item Context Independence: Lexical effects remain consistent across different sentence contexts.
\end{enumerate}


\subsection{Contextual Integration}
We create test cases that check how the model changes the sentiment of the raw words to the right meaning based on context. Our test suite includes test cases with the following sentiments: Medium intensity, Intensified swap, Simple negation, Intensified negation, Complex double negation, Domain context, Sarcasm, Intensity, Multiple intensifier, Scale variation.





\subsection{Hypotheses}
We test whether contextual integration follows the predicted layer specialization pattern:
\begin{enumerate}
    \item Middle Layer Concentration: Contextual effects peak in layers 4-8.
    \item Phenomenon Specificity: Different context types show distinct layer patterns. 
    \item Distributed Processing: Effects concentrate in specific layers rather than being distributed.
\end{enumerate}

\section{Data}
This section outlines the data generation methodology employed to evaluate the  sentiment processing hypothesis in GPT-2 mentioned in the earlier section.

\subsection{Lexical Detection Dataset:}


The dataset included 1,000 test cases across six types of contextual changes, each targeting different aspects of sentiment processing. I) Simple Negation cases examined basic polarity reversal through negation words (e.g., ``The movie was good'' vs. ``The movie was not good''). II)  Intensified Negation tested stronger negation patterns with adverbs (``The film was excellent'' vs. ``The film was definitely not excellent'').  III)  Sarcasm cases required the detection of situational incongruity (“Great, another meeting” with positive/negative contextual framing). IV)  Domain context examples tested how domain knowledge affects sentiment interpretation (``The horror movie was terrifying'' where `terrifying' is positive for horror but negative for other genres). V)  Intensification cases examined how modifiers amplify sentiment (``The meal was good'' vs. ``The meal was extremely good''). VI) Complex Double Negation tested sophisticated logical reasoning (``I don’t think it's not good'' requiring multiple negation resolution steps).

\subsection{Contextual Integration Dataset:}
The Contextual Integration Dataset comprises 8,000 carefully constructed test pairs designed to evaluate how GPT-2 processes context dependent sentiment modifications across 14 distinct phenomena. Each test pair consists of a clean sentence and a corrupted counterpart, differing only in specific contextual elements. The dataset systematically explores diverse contextual mechanisms: 
C1) Strong Positive: Substituted strong positive sentiment words with strong negative counterparts (e.g., `incredible' → `abysmal', `wonderful' → `horrible').
C2) Medium Intensity: Swapped medium-intensity positive words with medium-intensity negative words (e.g., `fine' → `bad', `enjoyable' → `unsatisfactory').
C3) Intensified Swap: Combined intensifier adverbs with opposite sentiment words (e.g., `utterly wonderful' → `utterly awful', `completely amazing' → `completely dreadful').
C4) Comparative Context: Changed comparative phrases and their sentiment outcomes (e.g., `better than expected, quite satisfying' → `worse than expected, quite mediocre').
C5) Simple Negation: Added or removed basic negation words to flip polarity (e.g., `nice' → `not nice', `are decent' → `aren't decent').
C6) Intensified Negation: Applied negation to intensified positive phrases (e.g., `was quite outstanding' → `wasn't quite outstanding', `very spectacular' → `wasn't very spectacular').
C7) Complex Double Negation: Used double negation patterns with contrasting outcomes (e.g., `wasn't bad at all, actually decent' → `wasn't good at all, actually disappointing').
C8) Domain Context: Changed domain context to alter sentiment interpretation of domain-specific words (e.g., `horror movie was haunting' [positive] → `romantic comedy was haunting' [negative]).
C9) Sarcasm: Modified intensity of sentiment words in sarcastic contexts to change perceived sentiment (e.g., `Perfect, amazing weather' [sarcastic/negative] → `Perfect, decent weather' [less negative]).
C10) Conditional vs Actual: Switched between conditional and actual statements to change sentiment (e.g., `would have been outstanding if not for' [negative] → `was outstanding despite' [positive]).
C11) Intensity Variation: Reduced or increased intensity modifiers while keeping base sentiment (e.g., `incredibly pleasant' → `a bit pleasant', both positive but different intensity).
C12) Multiple Intensifiers: Removed stacked intensifiers to reduce sentiment strength (e.g., `utterly very adequate' → `just adequate', positive → neutral).
C13) Intensity Flip: Changed strong intensifiers to weak/minimal intensifiers (e.g., `extremely spectacular' → `only slightly spectacular', positive → neutral).
C14) Scale Variation: Swapped sentiment words at different positions on the sentiment scale (e.g., `pleasant' [+3] → `horrible' [0], varying scale distances).


\section{Result}
\subsection{Lexical Detection}
To test our four part framework for lexical sentiment processing, we ran three experiments on early layers. The Lexical Sensitivity test checks if sentiment effects are strongest at word substitutions (Hypotheses 1–2). The Position Specificity test compares patching at sentiment vs. non-sentiment words (Hypothesis 3). The Context Independence test measures whether lexical effects stay stable across contexts (Hypothesis 4). Together, these activation patching experiments provide causal evidence for our framework beyond correlations.

\begin{figure}[h]
    \centering
    \includegraphics[width=\textwidth]{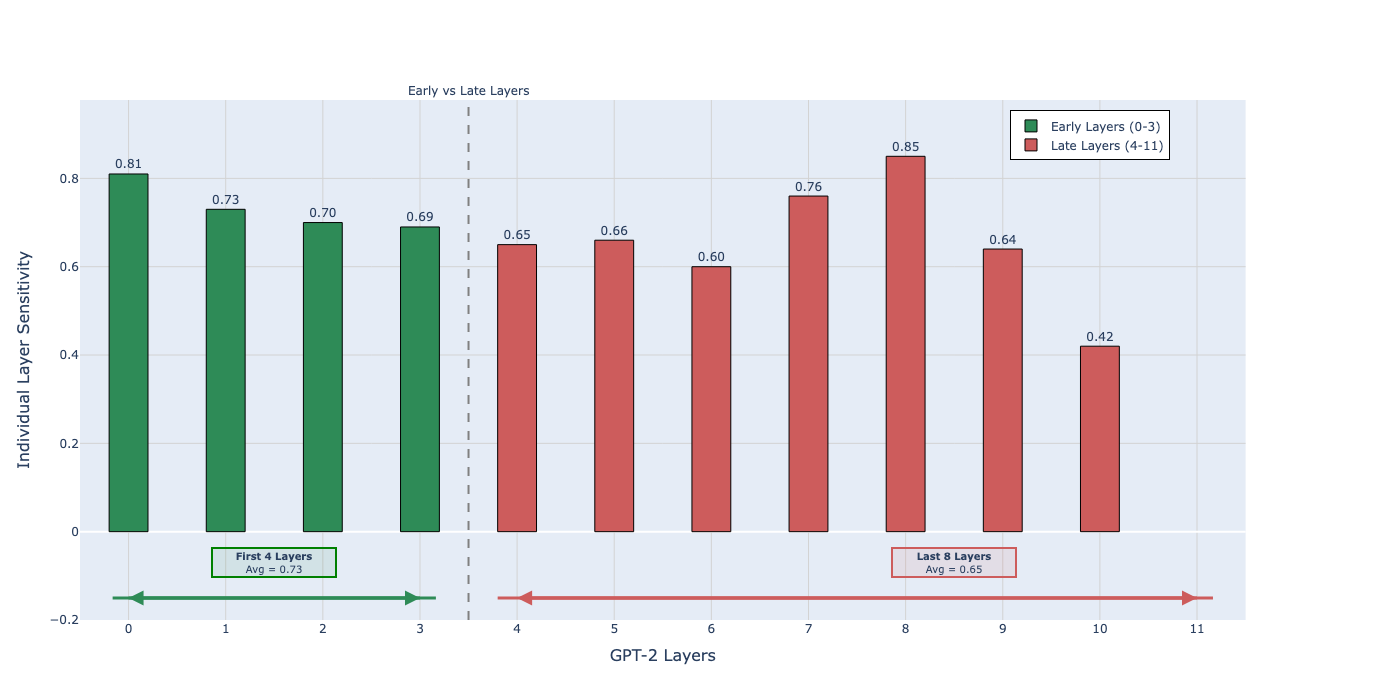}
    \caption{Lexical Sensitivity}
    \label{fig:layer_sensitivity}
\end{figure}

\subsubsection{Lexical Sensitivity }
For each sentence pair, we patched activations across all 12 GPT-2 layers. The clean sentence used a positive word, while the corrupted one used its negative counterpart. By replacing activations at target word positions, we measured each layer’s causal role in sentiment prediction. Position specificity was tested by comparing effects at sentiment vs. non-sentiment words, while context independence was measured by variation of these effects across different contexts. Figure \ref{fig:layer_sensitivity}, shows the lexical sensitivity in GPT-2 layers. Sensitivity quantifies how much a model's output (e.g., sentiment prediction) changes in response to interventions on internal components (e.g., neurons or attention heads). Bar heights show effect sizes from activation patching experiments. Average sensitivity of early layers Layer-0 to Layer-3(L$_0$-L$_3$) shows higher sensitivity to sentiment word substitutions, with L$_0$ exhibiting peak performance.

\begin{figure}[h]
    \centering
    \includegraphics[width=\textwidth]{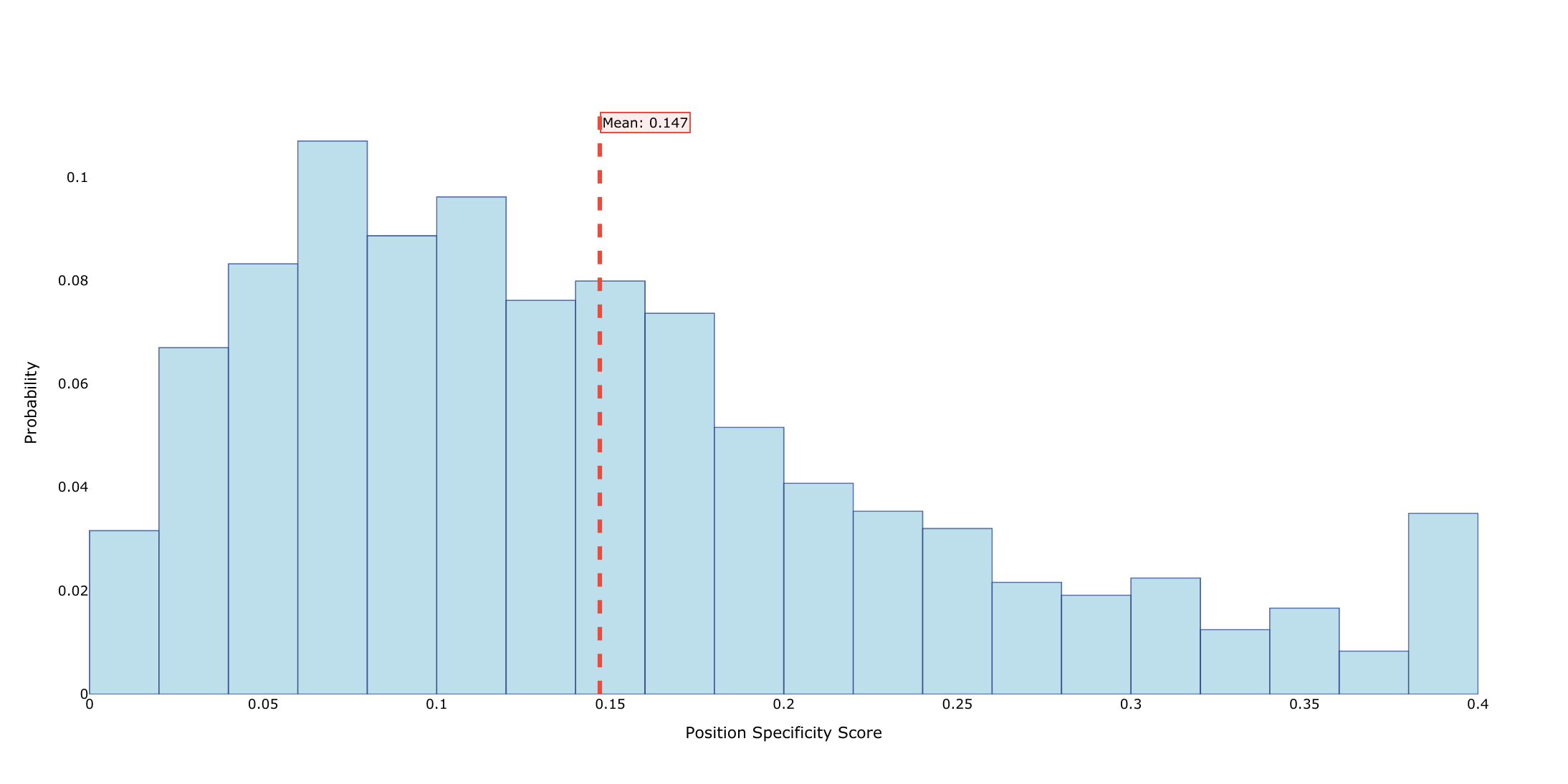}
    \caption{Position Specificity}
    \label{fig:position_specificity}
\end{figure}

\subsubsection{Position Specificity} 

Position Specificity Analysis tests whether GPT-2 detects lexical sentiment at precise word locations or through more diffuse sentence wide signals. We measure this by comparing the effect of activation patching at sentiment bearing words versus at non-sentiment words within the same sentence. Theory predicts that early layers should show strong word level sensitivity, since lexical sentiment features are tied directly to specific tokens. Figure \ref{fig:position_specificity}, shows our experiments confirm across 2000 test pairs, activation patching produced significantly stronger effects at sentiment word positions than at non target positions, with a mean specificity score of 0.147 (p < 0.001). This provides clear evidence that early layers, particularly Layer 1, localize sentiment information to specific token positions. In other words, GPT-2's lexical stage operates through targeted, word level detection rather than holistic sentence processing. This position specific encoding forms the foundation for later contextual integration stages, where sentiment must be adjusted through distributed processing to capture complex patterns such as negation or sarcasm.

\begin{figure}[htbp]
    \centering
    \includegraphics[width=\textwidth]{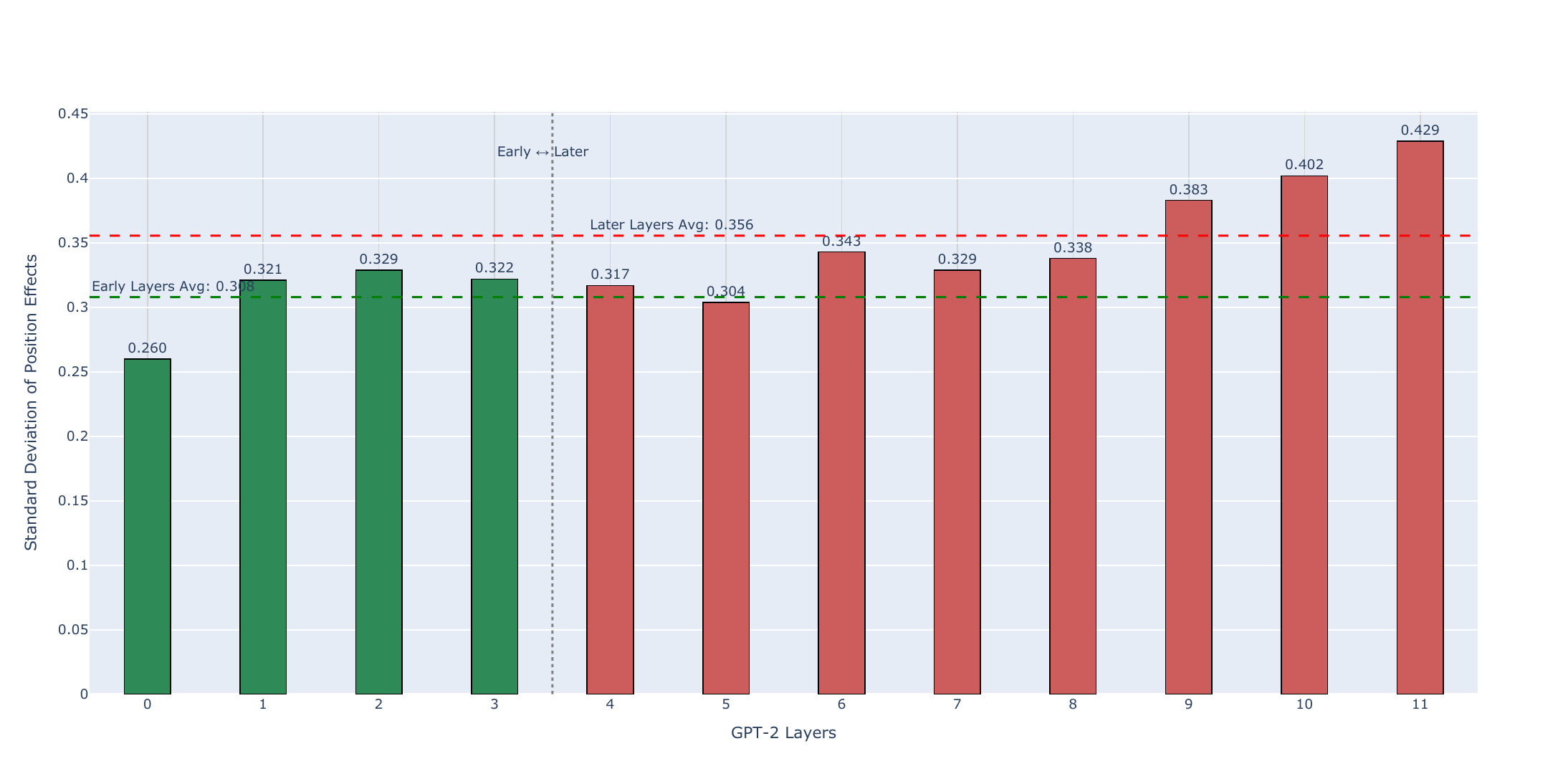}
    \caption{Context Independence of Sentiment Effects}
    \label{fig:Context_Independence}
\end{figure}

\subsubsection{Context Independence}

The context independence analysis tests whether the GPT-2 detection of lexical sentiment remains stable across different sentence contexts. We measure this by looking at how much the effect of sentiment words varies when they appear in different linguistic environments. If a layer is truly performing lexical processing, the effect of words like `wonderful' or `terrible' should remain consistent regardless of context. In contrast, context dependent processing should produce higher variability, since the same word may shift meaning depending on surrounding words. Figure \ref{fig:Context_Independence} shows early layers (L$_0$–L$_3$) shows  very low variability (mean = 0.038) in position effects, while later layers (L$_4$–L$_{11}$) show much higher variability (mean = 0.356). This confirms that early layers extract stable, context independent sentiment features, providing a reliable foundation for the rest of the network.

\subsubsection{Hypothesis Evaluation}
Lexical detection analysis shows that GPT-2's early layers reliably detect lexical sentiment. The results support all four hypotheses: (1) lexical sensitivity, (2) the early layers show the strongest sensitivity to sentiment words, (3) the effects are position specific, strongest at the locations of the sentiment words, and (4) the detection is context independent, with less variability than the later layers. Together, this confirms that GPT-2 encodes stable lexical sentiment signals early, forming the basis for later contextual integration.

\subsection{Contextual Integration}
To evaluate contextual integration, we tested three hypotheses in GPT-2: (1) Middle Layer Concentration, predicting peaks in  L$_4$-L$_8$; (2) Phenomenon Specificity, predicting distinct layer patterns for different contextual types; and (3) Distributed Processing, predicting effects spread across layers. Using controlled activation patching across all 12 layers, we measured the causal impact of specific contextual interventions on sentiment, isolating each phenomenon while keeping baseline conditions constant.

\subsubsection{Middle Layer Concentration}
\begin{figure}[htbp]
    \centering
    \includegraphics[width=\textwidth]{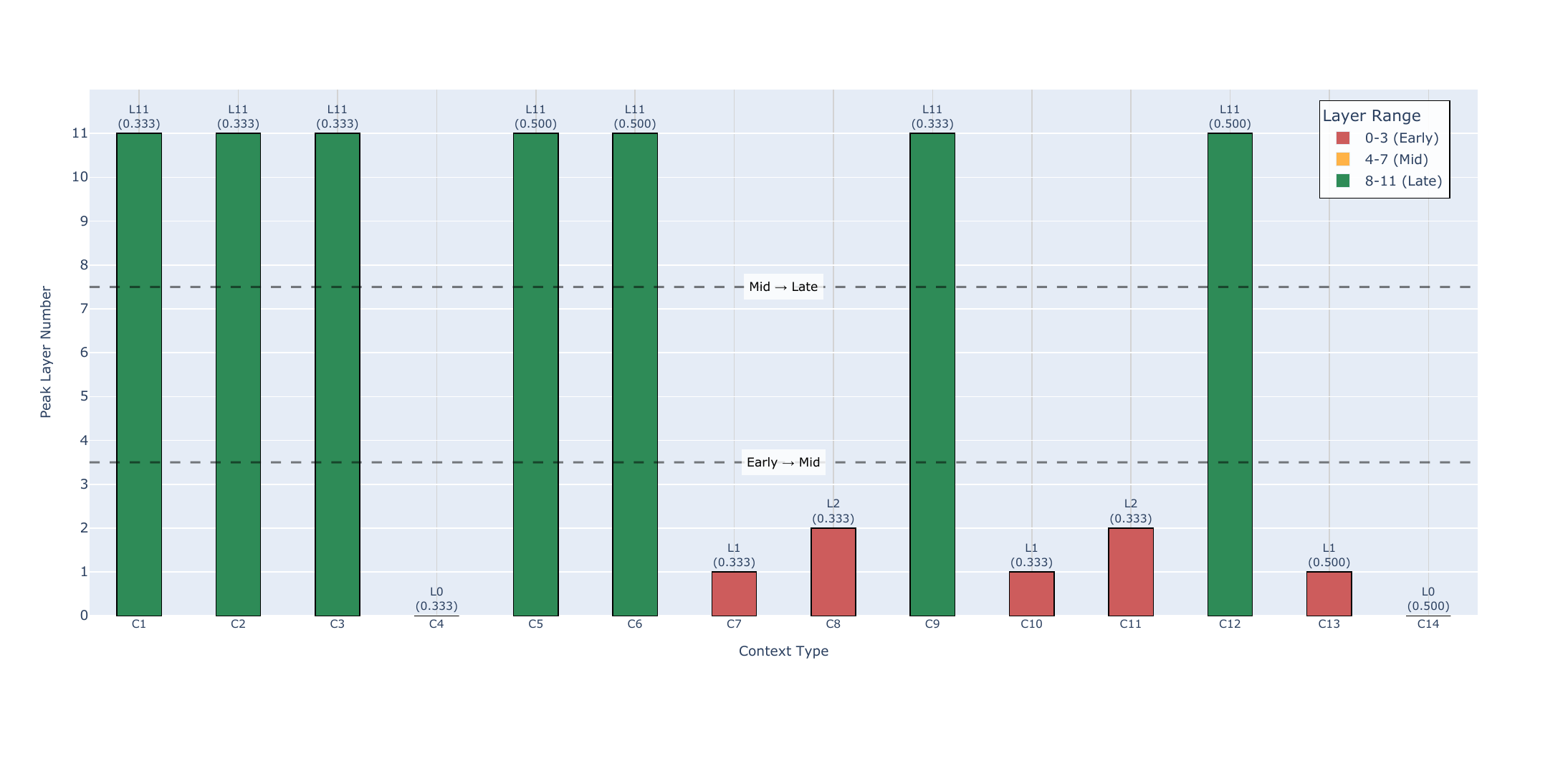}
    \caption{Peak Layer Distribution Across Context Types}
    \label{fig:peak_layer_distribution}
\end{figure}

The Middle Layer Concentration hypothesis predicted that contextual integration would peak in L$_4$-L$_8$, under the assumption that syntactic and semantic operations occur at intermediate depths of the network. Our experimental results, based on 8,000 test cases across 15 distinct contextual phenomena, contradicts this prediction.
Figure \ref{fig:peak_layer_distribution}, demonstrates that the network exhibits a bimodal distribution where phenomena cluster in either early L$_0$-L$_3$ or late layers (8-11), with no phenomena peaking in the predicted middle range L$_4$-L$_7$. Of the 15 contextual phenomena tested, 8 exhibit their

 strongest effects in L$_{11}$ (57\%), including strong positive contexts, medium intensity, intensified swap, simple negation, intensified negation, sarcasm, multiple intensifiers, and conditional vs actual contexts. The remaining 7 phenomena (43\%) peak in early layers: comparative context and scale variation (L$_{0}$), complex double negation, conditional vs actual, and intensity flip (L$_{1}$), and domain context and intensity variation (L$_{2}$). This bimodal pattern suggests fundamentally different processing strategies for different types of contextual modifications.


\subsubsection{Phenomenon Specificity}
The Phenomenon Specificity hypothesis predicted that different contextual phenomena would exhibit distinct layer-wise processing patterns, with each type of contextual modification recruiting specialized computational mechanisms at different network depths. Our experimental results decisively falsify this prediction, revealing instead remarkable convergence across semantically diverse phenomena. Of the 15 contextual types tested, 13 (87\%) share nearly identical top 3 contributing layers in the pattern 
[L$_{11}$, L$_{10}$, L$_{9}$] or [L$_{10}$, L$_{11}$, L$_{9}$]. Furthermore, 8 phenomena (53\%) peak at the exact same layer L$_{11}$. This convergence encompasses semantically diverse contextual modifications including simple negation, intensified negation, sarcasm, multiple intensifiers, and comparative contexts, all routing through the same late layer processing hub despite their different linguistic properties. Only domain context exhibits a genuinely distinct pattern, with both peak (L$_{2}$) and top-3 layers [L$_{2}$, L$_{3}$, L$_{4}$] concentrated in early-to-middle regions. This singular exception highlights what phenomenon specificity would look like if it existed systematically. The overwhelming convergence demonstrates that GPT-2 does not employ phenomenon-specific modules but instead processes most contextual modifications through a shared high-level semantic integration system.

\subsubsection{Distributed Processing}

\begin{figure}[htbp]
    \centering
    \includegraphics[width=\textwidth]{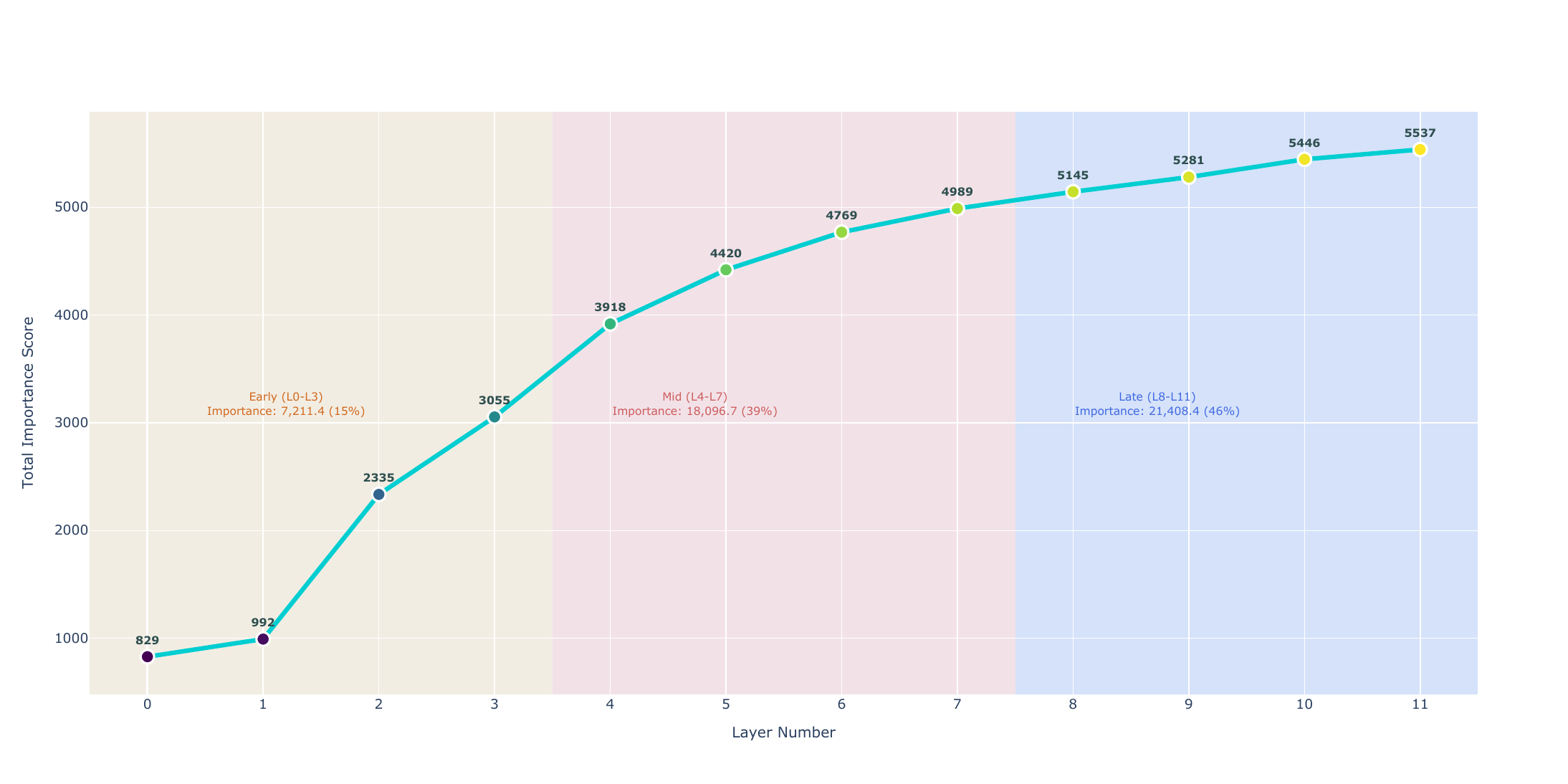}
    \caption{Layer Importance Gradient}
    \label{fig:layer_importance_gradient}
\end{figure}

The Distributed Processing hypothesis predicted that contextual effects would be spread across multiple layers rather than concentrated in specific regions of the network. Our results provide mixed evidence, revealing a more nuanced architecture than either pure distribution or strict concentration. The total layer importance analysis shows a clear gradient rather than a uniform distribution. Late layers (L$_8$-L$_{11}$) dominate with 46\% of all contextual processing, while mid-layers (L$_4$-L$_7$) contribute substantially 39\%, and early layers (L$_0$-L$_3$) account for only 15\%. The top five most important individual layers form a consecutive sequence from the network's upper regions: L$_{11}$, L$_{10}$, L$_{9}$, L$_{8}$, and L$_{7}$. Figure \ref{fig:layer_importance_gradient}, demonstrates that a monotonic decrease from late to early layers indicates a concentrated rather than distributed processing architecture. These findings largely falsify the Distributed Processing hypothesis in its strong form. Contextual integration is not uniformly distributed across all layers, but instead concentrates in a specific late layer region (L$_8$-L$_{11}$), with diminishing contributions from middle and early layers. 

\subsubsection{Hypothesis Evaluation}


Our systematic evaluation of the 8,000 case test dataset reveals that all three contextual integration hypotheses were falsified, though each provides distinct insights into GPT-2's processing architecture. The Middle Layer Concentration hypothesis failed as contextual phenomena exhibited 57\% late peaking and 43\% early peaking with zero phenomena peaking in the predicted middle layers (L$_4$-L$_7$), contradicting assumptions about intermediate layer semantic processing. The Phenomenon Specificity hypothesis was decisively rejected: 87\% of phenomena in our test cases share identical top 3 layers [L$_{11}$, L$_{10}$, L$_{9}$], demonstrating that GPT-2 routes semantically diverse contextual modifications negation, sarcasm, intensification through a unified late layer hub rather than specialized modules. Finally, the distributed processing hypothesis failed because the importance of layers in our dataset shows a 6.7 fold monotonic increase from L$_{0}$ (828.7) to L$_{1}$ (5,537.1), with late layers dominating 46\% of total processing weight. Together, these falsifications reveal an unexpected architecture within our experimental scope: GPT-2 employs phenomenon-agnostic late layer integration for most contextual reasoning, with only domain specific contexts (L$_{2}$, top 3: [L$_{2}$ , L$_{3}$, L$_{4}$]) representing a distinct early layer processing pathway. These findings, specific to our test dataset and methodology.

\section{Conclusion}
This study provides systematic causal validation of hierarchical sentiment processing in GPT-2 through mechanistic interpretability methods. We show that sentiment processing unfolds in a two stage: precise lexical detection in early layers followed by complex contextual integration concentrated in late layers, rather than in the predicted middle layers. All three contextual integration hypotheses  middle layer concentration, phenomenon specificity, and distributed processing were systematically falsified. These findings shows how rigorous activation patching can explain AI models  beyond correlational analysis to provide  causal insight into transformer computation. Future work should validate these patterns across diverse transformer architectures (BERT, RoBERTa, larger GPT models) to determine whether two-stage lexical-contextual processing represents a general architectural principle or remains specific to GPT-2's scale and training paradigm. Extension to fine-grained circuit-level analysis could identify the precise attention heads and MLP blocks responsible for lexical detection and contextual integration, moving beyond layer-wise analysis to map exact computational pathways within the transformer architecture.

\bibliographystyle{unsrtnat}  

\bibliography{main}

\end{document}